# Uncertainty-based Modulation for Lifelong Learning

Andrew Brna, Ryan Brown, Patrick Connolly, Stephen Simons, Renee Shimizu, Mario Aguilar-Simon[*]

*Intelligent Systems Laboratory, Teledyne Scientific, Research Triangle Park, NC, 27709, USA*

**Abstract**

The creation of machine learning algorithms for intelligent agents capable of continuous, lifelong learning is a critical objective for algorithms being deployed on real-life systems in dynamic environments. Here we present an algorithm inspired by neuromodulatory mechanisms in the human brain that integrates and expands upon Stephen Grossberg's ground-breaking Adaptive Resonance Theory proposals. Specifically, it builds on the concept of uncertainty, and employs a series of "neuromodulatory" mechanisms to enable continuous learning, including self-supervised and one-shot learning. Algorithm components were evaluated in a series of benchmark experiments that demonstrate stable learning without catastrophic forgetting. We also demonstrate the critical role of developing these systems in a closed-loop manner where the environment and the agent's behaviors constrain and guide the learning process. To this end, we integrated the algorithm into an embodied simulated drone agent. The experiments show that the algorithm is capable of continuous learning of new tasks and under changed conditions with high classification accuracy (> 94%) in a virtual environment, without catastrophic forgetting. The algorithm accepts high dimensional inputs from any state-of-the-art detection and feature extraction algorithms, making it a flexible addition to existing systems. We also describe future development efforts focused on imbuing the algorithm with mechanisms to seek out new knowledge as well as employ a broader range of neuromodulatory processes.

## 1. Introduction

> *A key issue leading to network models concerns how the behavior of individuals adapts successfully in real-time to constraints imposed by their environment*
> *– Grossberg (1987, pp. 24)*

State of the art machine learning algorithms, currently represented by deep neural networks, exhibit excellent performance when presented with test data similar to their training set, but fail when the test data differs from the training set in its statistical parameters (Azulay & Weiss, 2018; Neyshabur, Bhojanapalli, McAllester, & Srebro, 2017; Zech et al., 2018). Often these changes in statistical properties arise from changing conditions in the environment, or changes in the task itself. To properly classify these data, the network must be re-trained alongside the old training set to capture the features of the collective set. Simply retraining on the new task or dataset usually leads to catastrophic forgetting (French, 1999; Grossberg, 1980). This approach has at least two drawbacks: 1) the re-training necessarily requires the use of computational resources and time, and 2) there is no guarantee that the new training set captures the range of test data that will be encountered in the future. A more effective approach is one that addresses a solution to the *plasticity-stability dilemma* that Stephen Grossberg first articulated in 1980. It states that a system must remain plastic enough to learn important new information, while also maintaining stability in its memories for information it has already acquired.

---

[*] Corresponding author: Mario Aguilar-Simon, mario.aguilar@teledyne.com
Teledyne Scientific, LLC.

An elegant and robust solution is to develop life-long learning algorithms that employ novel architectures to engage in continual learning to support stable adaptation to new tasks, conditions and/or behaviors. Life-long learning architectures capable of stable learning could usher in an era of unprecedented performance, pervasiveness, and integration. Machine learning systems capable of continual learning will obviate the need for expensive data collection, labeling, and re-training that constrains state of the art systems today. Furthermore, life-long learning systems will increase robustness and safety by effectively permitting an autonomous system to learn to perform well in novel contexts and situations. As both fields of Neuroscience and Machine Learning advance, we envision continued synergy between insights gained from studying learning in neural systems, and new architectures that integrate those insights into novel machine learning systems.

In this article, we present a machine learning algorithm based on recent advances in neurophysiology and computational neuroscience that is capable of continual learning. The algorithm models the brain's modulation-based learning mechanisms (Grossberg, Palma, & Versace, 2016; Gu & Yakel, 2011; Hasselmo & McGaughy, 2004), utilizing metrics of uncertainty to guide lifelong learning in a system inspired by Stephen Grossberg's cognitive theory of Adaptive Resonance Theory (ART) (Grossberg, 1976; Grossberg, 2015; Grossberg et al., 2016). In a series of experiments, we show that our uncertainty-modulated learning (UML) algorithm is capable, in real time, of progressively adjusting its knowledge base, adapting to new conditions, and even incorporating never-before-seen objects, all without human intervention. A video demonstrating an embodied drone agent in a simulated environment utilizing the new algorithm is included in the additional material. The UML algorithm is to our knowledge the first to successfully demonstrate a solution to the stability-plasticity dilemma in a behaving agent. Its completion marks an important milestone in machine learning research, which owes its foundations to the work of Stephen Grossberg and colleagues.

## 2. Background

> *A fundamental problem of perception and learning concerns the characterization of how recognition categories emerge as a function of experience*
> *– Carpenter and Grossberg (1987b)(pp.240)*

### 2.1 Grossberg's foundational work

Stephen Grossberg has prescribed a systematic and effective approach to studying intelligent systems and deriving corresponding algorithms. His published work starting in the 1960s (Grossberg, 1967, 1968a, 1968b, 1969) set the stage for studying behavioral and neural mechanisms through principled mathematical models that captured critical mechanistic properties. While addressing complex phenomena, his Occam's razor approach to developing his computational models has proven to be extremely consequential to our understanding of seemingly disparate data and phenomena from a wide array of scientific disciplines including psychology and neuroscience. With the constant discovery of shortcomings in state-of-the-art machine learning algorithms (such as catastrophic forgetting), it is important that current practitioners in the field take a fresh new look at his body of work for guidance, if not inspiration.

There are six specific principles and contributions that Dr. Grossberg has articulated that served as a foundation to our UML algorithm. We list them here to provide the relevant background, and describe how they were adopted and implemented in our algorithm and experiments in their corresponding sections:

I. **Embodied development**: Early in his work, he suggested a theory that "approaches the problem of biological diversity by seeking *organizational principles* that have evolved in response to environmental pressures to which a surviving species must adapt" (Grossberg, 1978). The resulting analysis established that behavior is a critical anchoring for intelligent systems. He also demonstrated that a systematic decomposition of complex processes into the basic components is critical for an in-depth understanding of the behaviorally-relevant mechanisms that may be implemented in the brain.

II. **Emergent properties**: As part of this theory, he further outlined how "these organizational principles are translated into explicit neural networks that realize the principles in a minimal way. Once the properties of the minimal mechanisms are understood, variations of them can be more readily recognized and analyzed …[resulting in] a small number of robust principles and mechanisms that form a common substrate for coping with many tasks". In this and subsequent articles, Steve provided a roadmap for how to decompose complex phenomena to its simplest behaviors and how to analyze their integration. Additionally, he has demonstrated how a similarly systematic analysis must be follow to understand the emergent complex properties that arise from an integration of the basic components.

III. **Plasticity-stability dilemma**: To analyze necessary mechanisms for adaptive and robust behaviors in an embodied agent, he proposed a series of solutions to the plasticity-stability dilemma. He asked (Grossberg, 1982a), how is it that "internal representations can maintain themselves in a stable fashion against the erosive effects of behaviorally irrelevant environmental fluctuations, yet can nonetheless adapt rapidly in response to environmental fluctuations that are crucial to survival". Through a parsimonious process, he introduced a series of computational model components that provided critical functions for pattern learning, recognition and adaptation that eliminated the problem of catastrophic forgetting. It began with his 1967 paper where he described neural networks that can learn, simultaneously remember, and reproduce on-demand any number of spatiotemporal patterns (Grossberg, 1967). The paper introduced the complementary networks known as instar and outstar which provide mathematically sound models for discriminative and generative pattern learning that he integrated to demonstrate self-stabilized learning. Later work in collaboration with Gail Carpenter and other colleagues, formalized these mechanisms into the family of Adaptive Resonance Theory (ART) models that have led to successful algorithm implementations and applications (Gregory P Amis, Carpenter, Ersoy, & Grossberg, 2009; Cao, Grossberg, & Markowitz, 2011; Carpenter & Grossberg, 1987a, 1987b, 1987c; Carpenter & Grossberg, 1987a, 1987b; Carpenter & Grossberg, 1991; Carpenter, grossberg, & Reynolds, 1991; Carpenter, Grossberg, & Rosen, 1991; Grossberg, Govindarajan, Wyse, & Cohen, 2004; Palma, Grossberg, & Versace, 2012; Raizada & Grossberg, 2003).

IV. **Top-down**: One of the most important mechanisms associated with ART models is the interplay between bottom-up activation of pattern categories, or hypotheses, and the top-down learned priors for each hypothesis (Carpenter & Grossberg, 1987b; Grossberg, 1980; Grossberg, 1982b; Grossberg, 2013). During input processing, bottom-up connections activate a set of nodes in a higher layer associated with possible hypotheses for the (e.g., object) category of the input. During learning, the node(s) associated with the accepted hypothesis sample the input pattern in order to learn or refine a template for the pattern the hypothesis represents. Then, as a new input is analyzed and category hypothesis are generated, the template for the most highly activated hypotheses are compared with the input. The degree of match between these two patterns can trigger either a reset and a search for a different hypothesis, or a resonant state upon which weights are allowed to be modified. Notably, this mechanism elegantly prescribed how inferred hypotheses and network responses must first be checked against learned expectations before committing to an output. Such a mechanism is absent in neural network algorithms based on backpropagation and is the source of both their overconfidence and learning instability.

V. **Neuro-modulation**: The mechanisms for resonance and reset were further expanded and linked to neuro-modulation mechanisms in a series of papers describing the Synchronous Matching ART, SMART (Grossberg et al., 2016; Grossberg & Versace, 2008; Palma, Grossberg, et al., 2012; Palma, Versace, & Grossberg, 2012). Specifically, it was posited that neuro-modulation via the cholinergic system helps to establish and modulate the thresholds for the mismatch that triggers a reset or a resonant state during pattern analysis. This was well aligned with known evidence for the role of the cholinergic system in responding to mismatched expectations and facilitating synaptic plasticity(Saar, Grossman, & Barkai, 2001; Vogalis, Storm, & Lancaster, 2003; Yu-qiu, Shao-gang, Ya-ping, Zhi-qi, & Jun, 2016). The models began to explain the role of uncertainty in controlling pattern processing and learning. Specifically, how moderate levels of uncertainty-based modulation may trigger a test for a different hypothesis, and higher levels of uncertainty-based modulation may trigger one-shot learning of the input pattern (i.e., learn a new category).

VI. **Creation of explicit algorithms from neural mechanisms**: Finally, Grossberg's body of work is made even more relevant for the current state-of-the-art as it includes the development of specific algorithm solutions that capture many of the above computational principles. Among the most successful are both ART (Aguilar & Ross, 1994; Carpenter, 2013; Carpenter & Grossberg, 1987a; Carpenter & Grossberg, 1988; Carpenter, Grossberg, & Rosen, 1991) and ARTMAP (G. P. Amis & Carpenter, 2007; Carpenter, 2003, 2010; Carpenter, Grossberg, Markuzon, Reynolds, & Rosen, 1992; Carpenter, grossberg, & Reynolds, 1991). The former represent classes of unsupervised clustering and classification algorithms. While the latter represent classes of supervised learning algorithms. While these algorithms have not enjoyed the level of success that deep neural networks have recently achieved, they do represent concrete solutions for specific problems confronted by autonomous intelligent system. As a result, the machine learning community may benefit from considering whether any of these algorithms may serve an important supplemental role as components in the design of complex learning systems. Our uncertainty-modulated learning algorithm is an example of such an undertaking.

*2.2 Related work*

There has long been an understanding of the critical flaws and susceptibility for catastrophic forgetting inherent in connectionist networks that are based on backpropagation (French, 1999; Grossberg, 1982a; Ratcliff, 1990). Much work has been focused on ameliorating the problem. A recent example is elastic weight consolidation (Kirkpatrick et al., 2017), which proposes a mechanism for selective plasticity. While effective for some tasks, the mechanism is not sufficiently robust to ensure protection of prior knowledge. Furthermore, it is not clear whether such a mechanism would support long-term stability in a real-time behaving agent.

The interest for lifelong learning capabilities is closely linked to developing algorithms that address the stability-plasticity dilemma (Ruvolo & Eaton, 2013; Silver, Yang, & Li, 2013; Thrun, 1998). Progressive neural networks gained significant traction as a solution to the dilemma by proposing a mechanism for transfer learning that preserves prior tasks by carrying around parallel streams for each learned task (Rusu et al., 2016). While simple in its architecture, it is difficult to envision how such a solution would be scalable in a truly lifelong context. Furthermore, solving the plasticity-stability dilemma has to be accompanied with effective methods for self-supervised learning since autonomous system can have limited access to labeled data. The proposal by Parisi et al. (2017) provides an explicit mechanism for addressing the stability-plasticity dilemma for deep learning architectures that also suggests a mechanism for self-supervising the learning process. Their results suggest significant robustness in the case of missing label information though still remaining susceptible to catastrophic forgetting under a variety of context and task-performance conditions. Therefore, there is a need to consider an algorithm architecture that systematically incorporates a number of mechanisms for modulating learning in the presence of a variety of uncertainty conditions.

## 3. Algorithm

*...an interaction between two functionally complementary subsystems is needed to process expected and unexpected events*
*– Grossberg (1982b)*

### 3.1 Uncertainty and Neuromodulation

The learning algorithm used in this work expands upon principles of uncertainty pioneered by Grossberg's ART algorithm. In ART, uncertainty is best embodied in the vigilance parameter, which controls how similar a sample's bottom-up input needs to be to an existing top-down expectation to incorporate the sample into it. If the similarity metric does not pass the vigilance criteria, then the system is considered uncertain, and it engages a different learning mechanism in order to introduce the sample into the knowledge base. Our main contribution is to introduce specific mechanisms for measuring different types of uncertainty that an embodied agent needs to be able to track in order to perform robustly under a variety of conditions. Through an extensive review of analysis of evidence, we have identified a broad number of additional forms of uncertainty that we incorporate into a new uncertainty-modulated learning (UML) algorithm. Furthermore, we expand the mechanisms that trigger changes in the learning system so that new information is incorporated in a fully self-supervised manner.

Figure 1 illustrates the generalized form of the UML algorithm. The algorithm accepts high-dimensional representations of data, and it makes an initial decision/hypothesis using that data based on its own existing knowledge base. The hypothesis, the end point of traditional machine learning techniques, could take different forms based on the nature of the inputs; in this work it represents a classification, it could also represent a translation, a prediction, a diagnosis, or even a motor action. The uncertainty algorithm follows the initial hypothesis by calculating a set of uncertainty metrics, which represent sources of noise or confusion in the signals or decisions/hypotheses that may have an impact on the algorithm's output. Uncertainty can come from a variety of sources, including but not limited to the hypothesis, the internal representations of the existing knowledge base, the inputs themselves, and the conditions under which the inputs were received. Fig. 1 offers three such uncertainties.

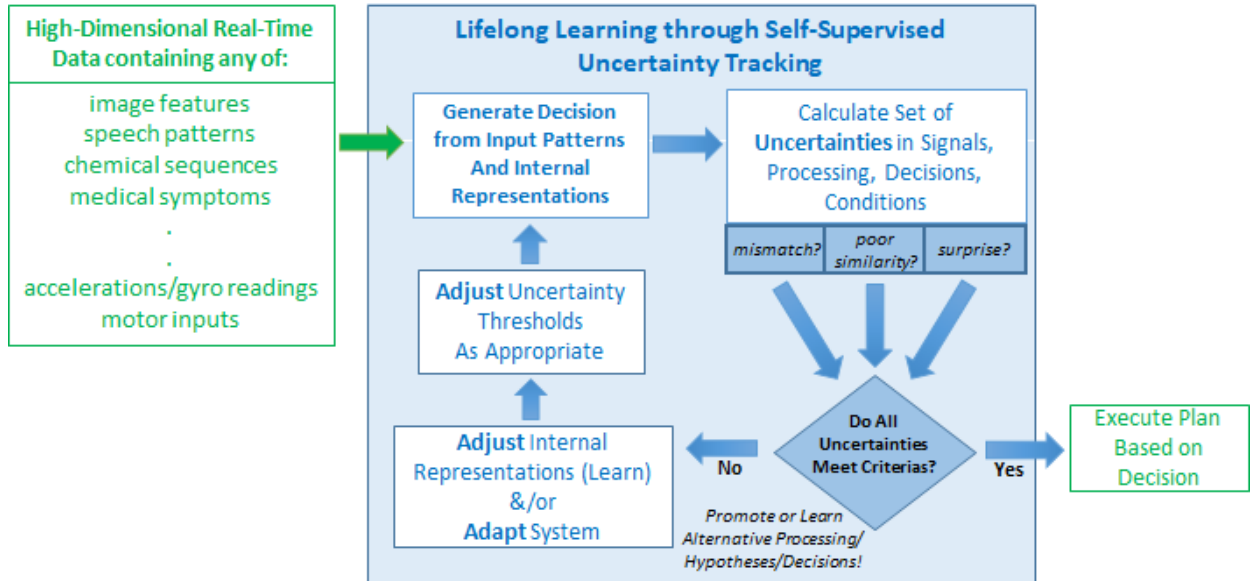

**Figure 1: Generalized Algorithm for the Uncertainty-Modulated Architecture**. Graphic shows the operation of the uncertainty architecture, which processes accepted high-dimensional inputs, calculates multiple types of calculated uncertainty, and compares them against internal criteria. Based on the comparison, the architecture either passes a conclusion to subsequent parts of the system, or it modulates internal values and repeats the process in that new state. This loop continues until all criteria are met.

Next, the uncertainty metrics are compared against internal criteria representing the algorithm's tolerance for each uncertainty type. If all uncertainty criteria are met, representing a confident hypothesis, the architecture finalizes its decision and passes it on to downstream components of the system. However, if any of the uncertainties do not pass criteria, then the algorithm changes, or modulates, its operations based on the specific uncertainty failure. Changes can be temporary, lasting only for one input, or permanent, changing the knowledge base for all subsequent inputs. Finally, the algorithm adjusts its criteria as appropriate for the failed uncertainties, and a new hypothesis is generated. This process repeats until the algorithm finds or develops a hypothesis that satisfies all uncertainty criteria.

Changes resulting from uncertainty are mediated by modulation mechanisms inspired by the acetylcholine neuromodulatory system (Grossberg et al., 2016; Yu & Dayan, 2005) and norepinephrine (Dayan & Yu, 2006). These neuromodulators can be released in response to expected uncertainty and surprise, triggering multiple effects on signal processing and neuroplasticity in multiple brain regions (Disney, Alasady, & Reynolds, 2014; Grossberg, 2017; Hangya, Ranade, Lorenc, & Kepecs, 2015; Kravitz, Saleem, Baker, Ungerleider, & Mishkin, 2013). Some of those effects induce temporary adaptations in cortical networks (Dasgupta, Seibt, & Beierlein, 2018), and others induce extended changes in hippocampal memory (Gu & Yakel, 2011). These neuromodulatory mechanisms form the basis by which the UML algorithm can incorporate new information into its knowledge base without destroying its prior knowledge.

### 3.2 Types of Uncertainty

The UML algorithm measures uncertainty with the principal goal of confidently adapting to the environment or task changes. Specifically, uncertainty allows the algorithm to monitor its performance against expectations and respond with the appropriate form of adaptation or response. UML currently evaluates five types of uncertainty: detection, category fit, similarity, relevance, and persistence. The exact sources of each type will vary among algorithm implementations, but as a whole they represent critical questions that a continual learning system must ask to determine when, what, and how it should learn. Detection relates to the uncertainty that a system has appropriately detected an object of interest. Fit reflects uncertainty in how closely the inputs match internal representations of the knowledge a system has already learned; ART examines this type of uncertainty to an extent through its vigilance parameter ρ and its "reset and search" and "match tracking" mechanisms. Similarity asks how well a sample relates to everything the system has already experienced; i.e., does this sample share any common ground with what has already been seen, or is it something entirely new? Relevance is uncertainty in the relationship between the learned information (e.g., objects it knows) and the current input. For instance, whether the input is related or similar enough to the classes of inputs it has previously learned. Lastly, persistence relates to temporal or observational uncertainty, as it examines consistency in both the knowledge to be learned and the system's understanding of that knowledge.

Since a principal mechanism for measuring uncertainty in our algorithm is to compare the inputs or feature activations against learned expectations, it can readily incorporate any machine learning method that can learn priors (e.g., Bayesian networks). Algorithm 1 gives the variant modeled after default ARTMAP 2 used in this work, which is designed to perform a classification task within a simulated environment (see sec. 5). Parameters related to Algorithm 1 are listed in Table 1, and rationale for ARTMAP's equations and parameters are described in the associated article (G. P. Amis & Carpenter, 2007). This variant examines five types of uncertainty, compares them to criteria $\Psi_i$, and triggers modulation to control processing flow accordingly. Uncertainty criteria $\Psi_i$ are given values at the start of operation but can be modulated during operation to promote adaptation under different processes.

## ALGORITHM 1

**Algorithm1: Uncertainty Algorithm with Modulation (default ARTMAP 2 variant)**

- Initialize uncertainty criteria $\Psi_{1-5}$ [see Table 1]
- For each frame, generate $d$ detections with features $a$ and objectness through object detector
- For detections $i = 1…d$, perform the following:
  1. Evaluate detection metrics against detection criteria $\Psi_1$
     1.1. If detection does not pass criteria, continue to next detection
  2. Evaluate complement-coded features $A=(a,a^c)$ against category fit criteria $\Psi_2$ using the following process
     2.1. for each network node $j = 1....C$, calculate node activation levels $T_j = |A \wedge w_j| + (1 - \alpha)(M - |w_j|)$
     2.2. collect activated node subset $\Lambda = \{\lambda = 1…C : T_\lambda > \alpha M\}$
     2.3. set winning node $J = \text{argmax}(T_\Lambda)$
     2.4. evaluate node activations $T_J$ against category fit criteria $\Psi_2$
        2.4.1. set vigilance $\rho$ = category fit criteria $\Psi_2$
        2.4.2. if $T_J$ does not pass vigilance $\rho$, remove $J$ from $\Lambda$ and return to step 2.3 [reset]
        2.4.3. if $T_J$ passes vigilance $\rho$, assign an initial label from node $J$ and attempt to learn
           - if supervisory label is provided & node J's label does not match,
             ○ modulate criteria $\Psi_2$ to reject label mismatch [match tracking]
             ○ remove $J$ from $\Lambda$ and return to step 2.3
           - otherwise, update node weights $w_J$ to incorporate detection $i$
             ○ node weights $w_J = \beta(A \wedge w_J) + (1 - \beta)w_J$
             ○ proceed to step 3
     2.5. if no nodes pass criteria $\Psi_2$, examine similarity of features $A$ to existing nodes
        2.5.1. for top nodes in original node subset $\Lambda$, calculate overlap( features $A$, node weights $w_j$)
        2.5.2. for first node $J$ that passes similarity criteria $\Psi_3$,
           - otherwise, update node weights $w_J$ to incorporate detection $i$
             ○ node weights $w_J = \beta(A \wedge w_J) + (1 - \beta)w_J$
           - proceed to step 3
     2.6. if no nodes pass criteria $\Psi_3$, create new category to hold novel sample
        2.6.1. create new label $N = n + 1$ [new class]
        2.6.2. create new node with label $= N$, $w = A$
  3. store current category in memory
  4. if number of detections for label N or physical object remains below relevance criteria $\Psi_4$ for too long,

4.1. remove label N or physical object from consideration and learning
5. if number of detections for label N or physical object passes persistence criteria $\Psi_5$ is sufficiently large,
   5.1. finalize category as hypothesis and request human-relevant label if desired
continue to next detection (*i++*)

**Table 1: Algorithm Parameters.** Table describing parameters for the uncertainty algorithm including mathematical notation, description of parameter, and range of possible values.

| Notation | Parameter | Possible Values |
|---|---|---|
| $\Psi_1$ | uncertainty criteria for detection | (0, 1) |
| $\Psi_2$ | uncertainty criteria for category fit | (0, 1) |
| $\Psi_3$ | uncertainty criteria for similarity | (0, 1) |
| $\Psi_4$ | uncertainty criteria for relevance | (0, 1) |
| $\Psi_5$ | uncertainty criteria for persistence | (0, 1) |
| $C$ | number of nodes in system | (0, ∞) |
| $n$ | number of labels known to the system | (0, ∞) |
| $a$ | feature vector | (0, 1) |
| $A$ | complement-coded feature vector | (0, 1) |
| $M$ | number of complement-coded features | -- |
| $T$ | node activation level | (0, 1) |
| $\Lambda$ | subset of activated nodes | (1, C) |
| $J$ | winning node | (1, C) |
| $w$ | node weights | (0, 1) |
| $\alpha$ | signal rule parameter | (1, C) |
| $\beta$ | learning fraction | (0, 1) |
| $\varepsilon$ | match tracking | (-1, 1) |
| $N$ | self-supervised label (class) | (1, *n*) |
| $H$ | final hypothesis | (-1, *n*) |

Each criteria $\Psi_i$ relates to a specific type of uncertainty which arises in a classification learning system, and their values reflect the type of uncertainty they are designed to address. Criteria $\Psi_1$ relate to detection confidence, and controls whether specific objects are considered further ("objectness", (Redmon & Farhadi, 2018)). Criteria $\Psi_2$ and $\Psi_3$ relate to uncertainties in the analysis of the inputs, specifically confidence of a classification or hypothesis given pre-existing knowledge. Finally, criteria $\Psi_4$ and $\Psi_5$ relate to uncertainty in the observations. The former establishing the relatedness of the object to previously learned objects and the latter establishing the permanence or consistency in the observation of any new objects.

The algorithm begins by creating or loading an ARTMAP-based network and setting the uncertainty criteria. A separate detector (see sec. 5.2.1) finds objects in a scene and returns multi-dimensional features and an initial hypotheses, which are then fed to the network one-at-a-time. The first stage of modulation subsequently occurs in step 1, as the detected object is examined against two uncertainty criteria and potentially rejected as poor quality or undesirable. In step 2, object instances in the environment are extracted from environmental conditions (see sec. 5.2.2), and a provided or extracted label is assigned to the object.

Steps 3-6.9 embody ARTMAP itself. Features are first complement-coded, and if no nodes are available in the network, the one is created using those features and a corresponding supervised/unsupervised label. ART then implements a "forward pass" on the network and generates a level of activation $T$ in each available node. As described in sec. 2, ARTMAP uses the node activations to find the closest approximation "winning" node $J$, and the fit $T_J$ of the input features to node $J$'s calculated weights is compared against the vigilance parameter $\rho$. The vigilance parameter $\rho$ is modulated by three of the uncertainty measures ($\Psi_2$, $\Psi_3$, and $\Psi_4$) to enable self-supervised learning under appropriate levels of uncertainty. $\Psi_5$ is used to determine if sufficient evidence (observations) have been collected for a newly acquired object (see sec. 4.3). If this criteria is met, then the system preserves the new object class and can make it available for analysis by a human operator who can assign a human-readable label.

During training, if node $J$ satisfies criterion $\Psi_{2\text{-}3}$, and the provided label matches node $J$'s label if supervised, then the network incorporates the input features into node $J$'s weights using ARTMAP equations (G. P. Amis & Carpenter, 2007). A node's weights can be considered the node's "internal representation" or template of the part of feature space it has learned to recognize. However, if the node $J$ does not pass criterion $\Psi_2$, ART's "reset and search" is engaged, and a new node is selected (or made) for evaluation. This allows the algorithm to modify the network's knowledge base intelligently by focusing attention on the information most relevant to the hypothesis being made. If the $\Psi_2$ criterion is passed, but node $J$'s label does not match the assigned label, then the algorithm performs match tracking, increasing the value of $\Psi_2$ before starting reset and search. In this way, the algorithm is self-supervised, adjusting its uncertainty criteria automatically. Both match tracking and reset and search constitute modulatory mechanisms.

Once the algorithm has determined an appropriate category for a sample (including "I don't know"), the algorithm places that category into a list containing like-values for a specific object over time in step 7. Categories in this list are replaced with non-detections (zeros) over time, so if the list does not contain enough recognitions of an object to pass criteria $\Psi_{4\text{-}5}$, then there is uncertainty of the object's permanence, and the current detection is rejected. If criterion $\Psi_4$ is passed, but the frequency of the most common hypothesis in the list is below criterion $\Psi_5$, then there is uncertainty in the algorithm's ability to form a hypothesis, and the detection is also rejected. Both rejection instances further exemplify modulation. The algorithm only continues to step 8 and outputs the final hypothesis for a detected object if all uncertainty criteria have been passed.

### 3.3 Novelty

The UML algorithm also includes the option of examining an input's novelty to determine if it constitutes a not-yet-seen category. Algorithm step 6.3.2 shows that during learning, if an input does not match any existing nodes, the algorithm will create a new category or class for that input. The new class is immediately available for further refinement and classification. In this way, the algorithm can add entirely new classes in an unsupervised manner. This function, which is inherent in ARTMAP, represents a form of modulation, but more importantly, it provides the algorithm with a form of one-shot learning.

Furthermore, modulatory mechanisms allow the UML algorithm to engage one-shot learning during interaction with the environment. If the algorithm returns an "I don't know" hypothesis at step 6.9 for a detection, and criterion $\Psi_3$ is passed, then the algorithm evaluates the novelty of the detection against the entirety of the available knowledge base (step 6.10). Using similar steps to those described in sec. 4.2, the fit $T_J$ for each node is evaluated against $\Psi_2$. If a well-matched prior (node) is found, then the algorithm evaluates its location in the feature space relative to the input features. If this distance is within criterion $\Psi_3$, then the algorithm incorporates the detection into the best matched node for a previously-learned category. However, if a good match is not found, then the input is considered novel but relevant due to feature overlap with prior learned categories per criteria $\Psi_4$. In this case, one-shot learning is engaged, and a new class is formed with a randomly-assigned label. These processes combined give the uncertainty algorithm powerful lifelong learning capabilities, and we tested their functionality within this work.

## 4. Methods

### 4.1 Simulation Environment and Data Collection

Experiments run with the learning architecture utilized Microsoft AirSim (Shah, Dey, Lovett, & Kapoor, 2017) for data collection and testing. A separate autonomous vehicle and an aerial drone were used for data collection within a custom-designed urban environment. Each vehicle generated a standard RGB camera image (the "scene image"), and an image with segmented objects labeled by color (the "segmentation image"). Our custom urban environment, shown in Figure 2, featured a central roundabout and multiple outlying intersections; a mountain range was placed around the outside of the city to provide a finite end to the available camera range. Three of the intersections were populated with objects from one of three operating sets as described in sec. 6 (Sets A, B, and C), and 608x608 scene and segmentation images were collected from each for offline analysis by driving the camera-enabled vehicles through and around the intersections.

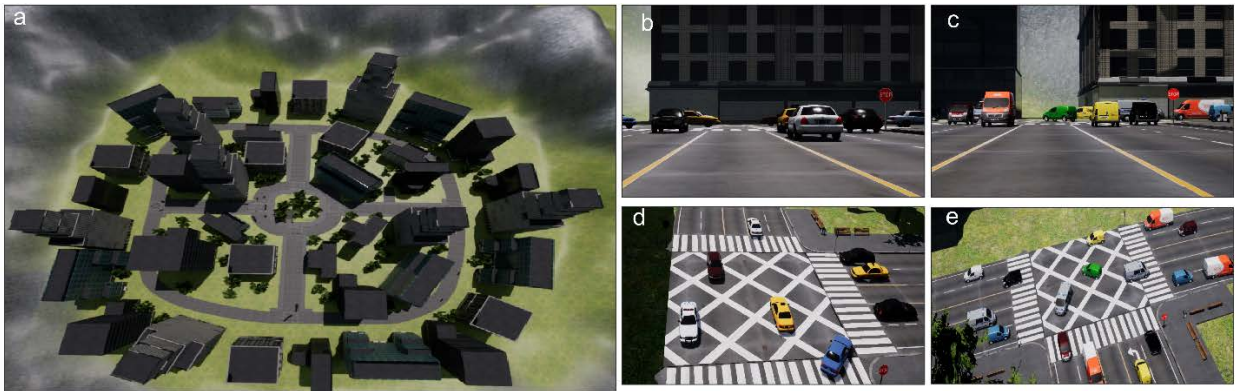

**Figure 2: AirSim Environment with Collectable Camera Images**. Graphic presents an example view of AirSim during development and during data collection. (a) shows an overhead view of the custom AirSim environment built to test the learning capabilities of the uncertainty architecture. (b,d) and (c,e) show the paired ground and aerial images used in data collection.

The images collected by the camera-enabled vehicles were split into two categories: ground data and aerial data. The ground data set included all images captured by the autonomous vehicle, which reflected a view approximately 1-2m off the ground. The aerial data set included all images captured by the drone, and these data ranged from approximately 8-25m off the ground, with a median height of 20m. For the learning experiments detailed in sec. 5.1, ground truth labels were generated by matching the bounding boxes returned by the detection system as described in sec. 5.2.1 with the object labels provided by the AirSim segmentation images. The labeled data were then split to create training and testing samples.

### 4.2 System Overview

The continuous learning architecture consists of three components: a detection subsystem, a spatial attention subsystem, and a classification subsystem. The detection and classification subsystems were used together for the learning experiments described in sec. 5.1, while all three subsystems were used in concert during the embodied system experiment described in sec. 5.2.

#### 4.2.1 Detection Subsystem

The first stage of the system assembly consisted of a detection subsystem designed to locate objects within the AirSim scene images and extract features from them. This subsystem used a version of the YOLOv3 algorithm (Redmon & Farhadi, 2018). Importantly, no additional training was done on the YOLO network. All YOLO model weights were pre-loaded (based on the COCO dataset, (Lin et al., 2014)) and remained constant for the duration of the experiments. Learning instead took place in the classification subsystem described in sec. 5.2.3. YOLO detected objects and their corresponding feature activations in the last convolutional layer were relayed to the classification subsystem.

#### 4.2.2 Spatial Attention Subsystem

The second stage of the system assembly was a spatial attention system designed to keep track of objects for the drone learning agent. The spatial attention subsystem served a similar function to the posterior parietal cortex in the mammalian visual system (Gottlieb, Kusunoki, & Goldberg, 1998), retaining the location of behaviorally-relevant objects as they entered and left the agent's field of view. This module supported continual learning by keeping track of objects for self-supervision. The spatial attention subsystem was implemented in Julia 1.1, a relatively-new dynamic computing language designed for both ease of use and speed of execution (Bezanson, Karpinski, Shah, & Edelman, 2012). This subsystem was used only during the fully-integrated system experiment described in sec. 5.2.

#### 4.2.3 Classification Subsystem

The classification subsystem contained a functionalized version of the uncertainty architecture detailed in sec. 2 and coded in Julia. The subsystem accepted features of detections with desired YOLO classifications and generated new classifications for analysis. Uncertainty criteria $\Psi_1$, $\Psi_4$, and $\Psi_7$ were manipulated to control operational states as desired during experimentation, including multiple training and testing modes. Object labels from the spatial attention subsystem were added as additional inputs during supervised training states, and only internally-available labels were used otherwise. Single-detection outputs stored over the latest ten frames were used to generate classes for display, but not learning, as described in sec. 2. The list did not store samples during the learning experiments described in sec. 5.1, and performance was evaluated with outputs directly from the classification subsystem ($H_N$, beginning of algorithm step 7). Uncertainty criterion $\Psi_3$ was changed during the fully-integrated system experiment to increase detection subsystem stringency as described in sec. 5.2.1.

### 5. Experiments and Results

*One begins with behavioral data because the brain has evolved in order to achieve behavioral success. Any theory that hopes to link brain to behavior thus needs to discover the computational level on which brain dynamics control behavioral success. Behavioral data provide a theorist with invaluable clues about the functional problems that the brain is trying to solve.*
*- Grossberg, 2006[†]*

Our experiments centered on a scenario with a recognition system trained with a ground-based autonomous vehicle that is subsequently ported to a flying drone. Such a transfer would typically suffer significant decreases in performance and require additional training to function properly. To show that our learning architecture is capable of adapting to such a dramatic shift in conditions, we conducted three experiments of increasing difficultly to examine the algorithm's continual learning and one-shot learning capabilities. Following these experiments, we created a continual learning experiment using the fully-integrated learning system in our simulated environment.

---

[†] *From personal monograph written in 2006, http://sites.bu.edu/steveg/files/2016/06/GrossbergInterests.pdf*
Teledyne Scientific, LLC.

### 5.1 Learning Experiments

The continual learning experiments examined the ability of the uncertainty algorithm to incorporate new information without losing its understanding of old information. Standard learning architectures are bound by their initial training data and augmentations, and they often are inflexible to use in new domains, requiring retraining from scratch to gain new functionalities. Table 2 illustrates this problem with a state-of-the-art YOLOv3 algorithm trained primarily on ground-based (COCO) imagery (Lin et al., 2014); while the algorithm is able to classify objects appropriately from a view matching its training set, its performance suffers dramatically when shown the same objects from the air. Our first experiment aimed to show that our uncertainty algorithm could learn to perform this task without retraining from scratch and without catastrophic forgetting.

**Table 2: YOLOv3 Performance from Ground and Aerial Views**. Table shows an analysis of YOLOv3's outputs from 11 scenes from either the ground or the air. On its own, a COCO-trained YOLOv3 is able to detect and identify objects from the ground with high accuracy. However, without the use of a continual learning mechanism, its performance drops severely when viewing the same scenes from the air. Increased object counts in the air result from an increased field of view.

| Object Type | Ground View Correctly Identified | Total | Aerial View Correctly Identified | Total |
|---|---|---|---|---|
| Car | 41 | 50 | 24 | 63 |
| Bus | 1 | 2 | 0 | 3 |
| Truck | 2 | 2 | 0 | 4 |
| Traffic Light | 14 | 17 | 0 | 24 |
| Person | 8 | 11 | 3 | 22 |
| Bench | 3 | 3 | 0 | 2 |
| Stop Sign | 1 | 3 | 0 | 3 |
| Total | 70 | 88 | 27 | 121 |
| % ID | **79.5%** | | **22.3%** | |

In our initial experiments, our data were split into three classes. Set A, contained a variety of sedans. Set B contained a variety of vans, with similar features to the sedans of Set A. The third class, Set O, contained an assortment of other objects relevant to traffic but resembling neither Set A nor Set B (e.g., park benches). Objects from Set A and Set B were placed in separate intersections with Set O objects interspersed throughout. As described in sec. 4.1, samples were collected for each set from both the ground and from the air. Table 3 lists the types of models included and total number of samples in each set; set C objects were not used in the continual learning experiments and are discussed in sec. 5.1.2.

**Table 3: Description of Object Sets**. Table gives AirSim models comprising Sets A, B, O, and C with associated sample counts in the ground and aerial data sets. Sets A, B and O were utilized the continual learning experiments, while Set C was reserved for one-shot experiments.

|  | Set A: Sedans | Set B: Vans | Set O: Other | Set C: Fire Trucks |
|---|---|---|---|---|
| Model | sedans, taxis, and police cars | mail vans, delivery vans, and police vans | stop lights, stop signs, crosswalk signs, fire hydrants, trash cans, and benches | fire trucks |
| Ground # | 677 | 1097 | 316 | -- |
| Aerial # | 645 | 1390 | 75 | 298 |

### 5.1.1 Initial Domain Transfer with Continual Learning

To represent the initial, ground-based machine learning system, the continual learning algorithm was first trained on all available ground-based imagery, including samples from Sets A, B, and O. Samples were shown to the system sequentially, in random order, and with class labels included, allowing the system to adjust its internal representations of the data as necessary to map each sample to the appropriate class. When successively tested on the same training data, the system correctly classified samples with 96.3% accuracy, confirming that the system was able to learn the initial dataset.

Following ground training, the transfer of the learning system to an airborne drone was simulated by exposing the learning architecture to aerial imagery of Sets A and O. 70% of the available aerial samples were used for training, with the remaining samples reserved as a test set. As expected, the initial performance on the aerial test samples prior to continual learning was poor, with only 26.7% of samples being correctly classified. Subsequently, the system was trained with the aerial training set. As with the ground dataset, samples were trained sequentially, in random order, with class labels included, and the system added new information to its knowledge base as well as modified previous knowledge. To emphasize, no ground samples were included in this new round of training.

In pre-existing machine learning architectures, sequential learning with samples from a different dataset would be expected to rapidly trigger catastrophic forgetting; performance on the new dataset would rise with additional training, but performance on the old dataset would fall rapidly. However, this was not the case with the uncertainty algorithm. Figure 3 and Table 4 show the results of incremental training, wherein the system was re-tested with the aerial test set as well as with the original ground data following exposure to certain percentages of the aerial training data. Over the course of training, performance on the aerial test set rapidly rose, reaching a plateau at 93.0%, reflecting that the system was able to learn the new dataset. However, not only did performance on the original ground data not fall, but it actually improved to 96.7%. This suggests that incorporation of aerial information sharpened the internal representation of the ground-based information as well. These results strongly indicate that the uncertainty algorithm is capable of continual learning without catastrophic forgetting.

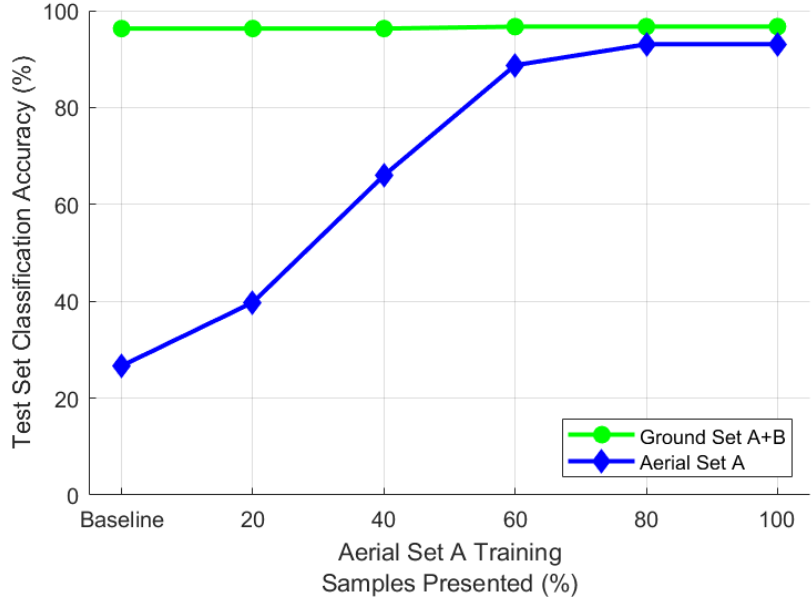

**Figure 3: Continual Learning with Set A Objects**. Graphic gives performance of the learning system on ground and aerial test sets over the course of training on aerial views of Set A (and Set O). As the learning system is exposed to progressively more samples from the training set, performance with aerial test data improves, while performance on ground data remains at the baseline of 96%.

**Table 4: Change in Performances Following Continual Learning with Set A Objects**. Table lists system classification performances on ground data and aerial test Set A before and after exposure to aerial views of Set A (and O) objects. Note that performance on Set A improves dramatically, and performance on ground data does not drop.

|  | **Ground Set A+B** | **Aerial Set A** |
|---|---|---|
| **Initial Performance** | 96.3% | 26.6% |
| **After Aerial A Training** | 96.7% | 93.0% |

In a different version of this experiment, we examined how learning information in sequence could improve the domain transfer. Using a separate collection of objects from Set A, the algorithm trained on ground-based imagery was trained sequentially at heights ranging from approximately ground level up to 20m. Its classification performance was then evaluated using a different set of test images containing similar views from heights up to 30m. Figure 4 shows that when trained only on ground views of the objects, classification performance dropped monotonically with increasing height, matching expectations. Subsequently training at a single height increased overall performance with the greatest improvement occurring at the trained height. In contrast, the best overall performance was obtained when the architecture was allowed to learn continuously from multiple heights. Performance after continual learning at two different heights was better at all locations than any single training height, including at points well outside the training ranges. These results suggested that continual learning was crucial for improved generalization overall, and they informed the structure of our embodied agent experiment.

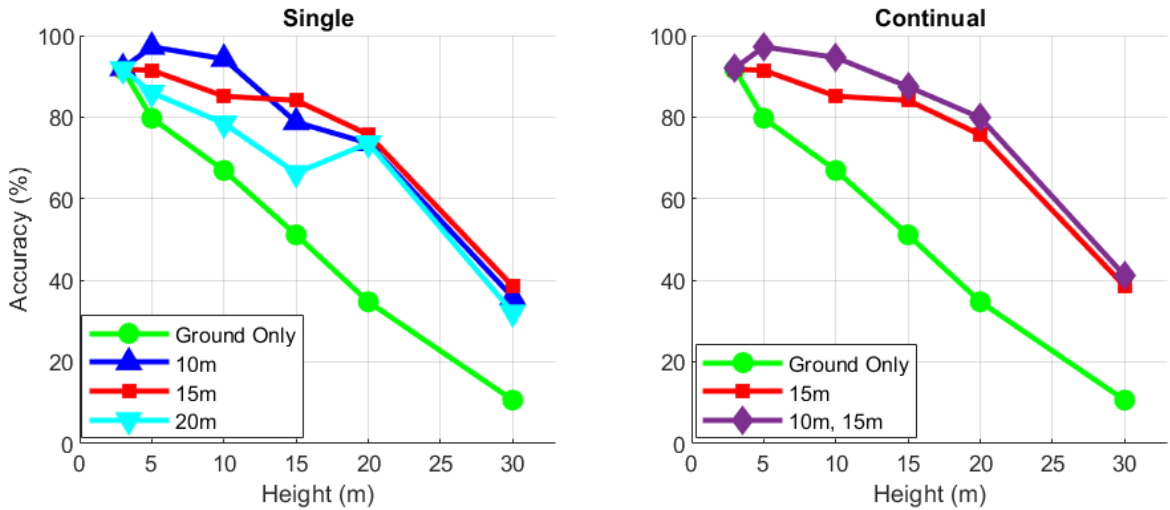

**Figure 4:** Continual Learning from Multiple Vantage Points. Left plot illustrates the effects of allowing a ground-based system to train on the ground imagery plus a single additional height. Each plot line represents the same starting system subsequently trained on imagery from the corresponding height. In all cases, additional training improved performance at all tested heights (as compared to baseline as indicted by the green plot line), but the greatest increase in performance was for the specific trained height. Right plot illustrates the effects of training on multiple heights sequentially. A system allowed to learn continually from two heights (purple plot line) out-performed the system trained at the single specific height.

### 5.1.2 Maintaining Class Boundaries in New Domains with Continual Learning

The first continual learning experiment effectively investigated the uncertainty algorithm's ability to transfer domains, but the second experiment more closely examined how well it could adjust the features to which it attended. In the second experiment, the learning system was subsequently exposed to aerial views of Set B. As the system had already been trained on both ground and aerial views of Sets A and O, but only ground views of Set B, it would be expected that the system would need to modify both its representations of Sets A and B as well as its understanding of aerial and ground views in order to perform well on the task. Similarly to the first experiment, 50% of the aerial views of Set B were set aside as a training set, with the remainder being used for testing. Training was done incrementally, and the ground training set, aerial test Set A, and the aerial test Set B were evaluated at regular intervals. No ground samples nor Set A samples were used for additional training.

Testing with the aerial views of Set B immediately following the initial ground training showed an overall performance of 9.1%, and that value did not change over the course of the first continual learning experiment. Included in Figure 5, this result suggested that while some information from the ground-based views was utilized in recognizing Set B from the air, training on aerial views of Set A had no meaningful impact that data set. By contrast, incremental training on aerial views of Set B had an immediate impact on performance, with classification rates for aerial views of Set B objects climbing rapidly to 94.1%. Performance on the initial ground training set remained at 96.7%, while aerial test Set A dropped by less than one percentage point. In short, the uncertainty algorithm successfully learned how to recognize old targets from a new view, without meaningfully degrading its understanding of other targets with similar traits. These results, also shown in Table 5, again strongly indicate that the UML algorithm is able to improve its function with new information without losing its previous functions, and it is able to do so in a progressive, continuous manner.

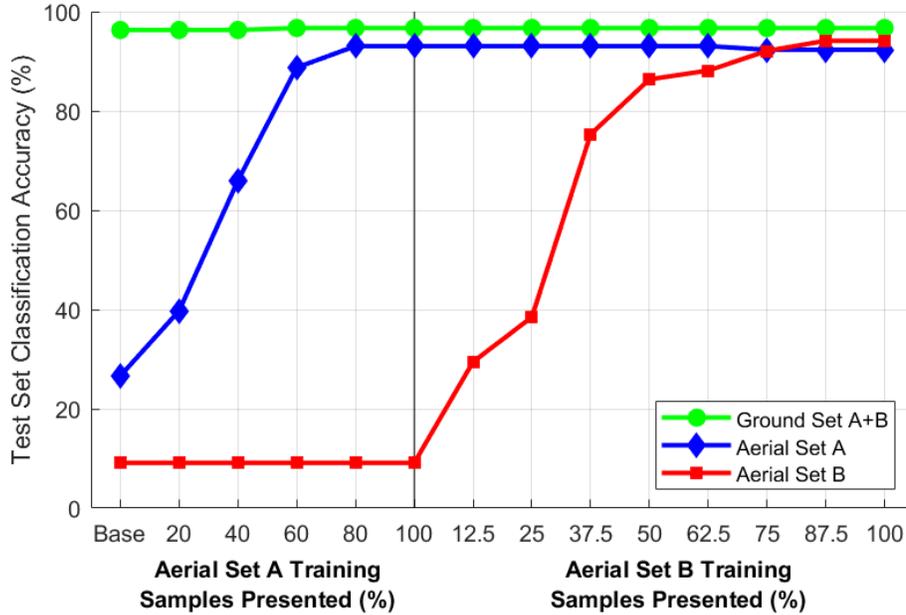

**Figure 5: Continual Learning with Set B Objects.** Graphic gives performance of the learning system on ground and aerial test sets over the course of sequential training on aerial views of Set A (and Set O) followed by Set B. The left half of the figure is the same as Fig. 4, with performance on aerial test Set B added. During training on Set A, performance on Set B is unaffected. The right half of the figure shows subsequent training on Set B. As the learning system learns with Set B, Set B performance quickly improves, finally reaching similar performance levels as the ground and Set A test data. Meanwhile, ground and Set A test performance remains virtually unchanged, demonstrating continual learning does not cause catastrophic forgetting.

**Table 5: Change in Performances Following Continual Learning with Set B Objects.** Table lists system classification performances on ground data, aerial test Set A, and aerial test Set B before and after exposure to aerial views of Set A (and O) objects followed by exposure to Set B objects. Note that performance on Set B rises to match previous test sets, and performance on other sets remains high.

|  | **Ground Set A+B** | **Aerial Set A** | **Aerial Set B** |
|---|---|---|---|
| **Initial Performance** | 96.3% | 26.6% | 9.1% |
| **After Aerial A Training** | 96.7% | 93.0% | 9.1% |
| **After Aerial B Training** | **96.7%** | **92.3%** | **94.1%** |

### 5.1.3 One-Shot Learning Experiment

The third learning experiment was designed to test the uncertainty architecture's ability to perform self-supervised one-shot learning. Traditional machine learning techniques require a massive number of samples (and retraining on all data) to learn to recognize a new class of objects (Li, Fergus, & Perona, 2006); in practice, this can be a significant problem, as the availability of such a pool of data is not guaranteed. Therefore, one-shot learning embodies the challenge that a system can learn to recognize, without external supervision, a new category of information from as few as one sample.

For the one-shot learning experiment, we added a new class of objects: firetrucks (Set C; Figure 6, left). This vehicle class was chosen as it is both similar to the other objects (e.g., they are all vehicles, have wheels, etc.) and significantly different to the other vehicles (e.g., size, number of wheels, structure, etc.) This is confirmed by projecting the YOLO activations via t-SNE (van der Maaten & Hinton, 2008) (see Figure 6) where the distribution of the aerial views of sedans (A) and vans (B) overlap significantly and those of the firetrucks (C) form a unique and separate cluster. This, along with their relevance to the task simulated in the scenario (i.e., counting vehicles from the air), made the firetrucks ideal candidates for this one-shot study.

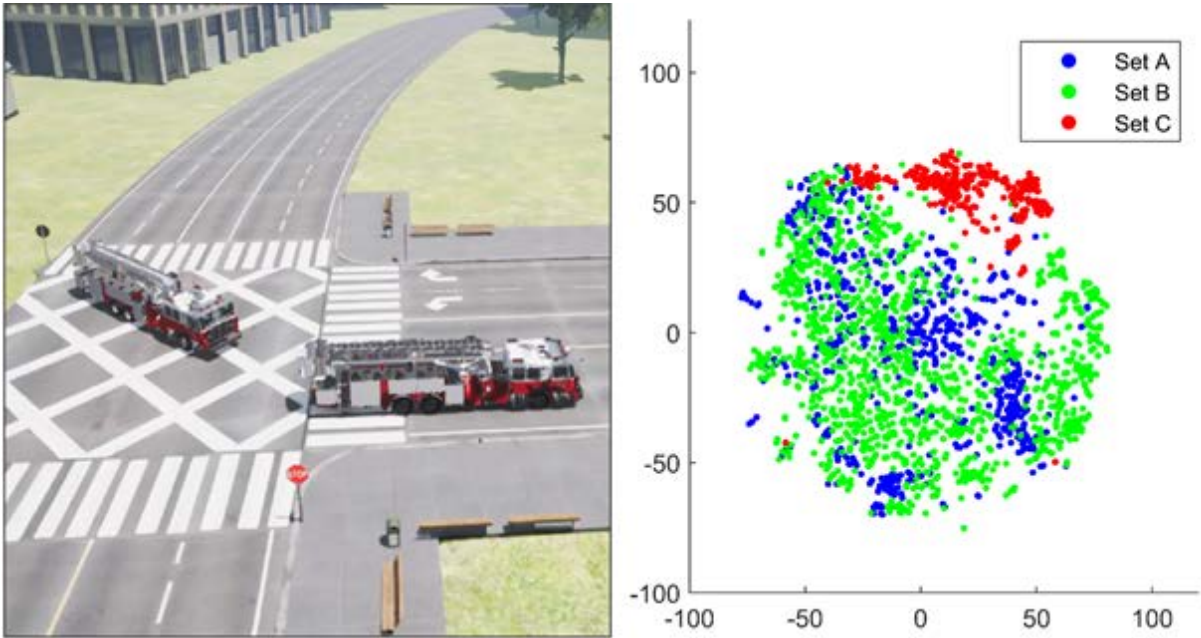

**Figure 6: Set Clustering with t-SNE**. Graphic demonstrates difference of Set C objects from Sets A and B objects when viewed from the air. Sets A and B objects overlap significantly, signifying a strong similarity among their features. However, Set C objects cluster away from Sets A and B, signifying a different distribution of feature values. This difference is detectable by the uncertainty algorithm, and drives the creation of new categories while engaged in self-supervised learning.

As with the previous two experiments, a series of aerial samples of Set C objects were collected in AirSim as described in sec. 5.1, and 10% of them, or 29 samples, were utilized as a training set, with the rest being reserved for testing. The total count of Set C samples collected is given in Table 2. The trained system from the end of the second learning experiment (aerial Set B) was used as the starting point for this experiment; note that to this point, the learning system had never been exposed to Set C objects, even from the ground. To promote self-supervised learning, an internal variable controlling the system's tolerance for uncertainty was reduced, and samples in the aerial training Set C were stripped of their classification labels. The learning system was subsequently shown the aerial training data for Set C sequentially, in randomized order, but without any supervisory signal to indicate the novelty of the Set C objects. Following exposure to the training set, the system's uncertainty tolerance was returned to its previous value to effectively disable further learning.

In response to its uncertainty of classification on Set C objects, the unsupervised learning system created multiple self-labeled classes, recognizing that the samples it was presented with did not match closely enough with anything in its pre-existing knowledge base. The new classes consisted of as few as one image or view apiece, but in several instances views were grouped with one another to make higher-confidence, conglomerate categories. In multiple cases, three or more views of the object were grouped together into new classes. This suggests that the groupings were formed based on similarities among samples with the same view.

### 5.1.4 Human-in-the-Loop Label Assignment

The self-supervised classes created by the uncertainty algorithm are internally consistent— a new sample whose features are within a certain range of values will be identified as a member of that class, and if learning is engaged, the representation of that class and only that class will be modified. These traits define one-shot learning. However, in the example of an autonomous agent, how should the agent treat the new classes it creates? Should they be avoided, as with a vehicle, or should they signal a change in operation, as with a stop sign? Such decisions may require the assignment of human-relevant labels, which can then be incorporated into an agent's behavioral controls. Interactive machine learning (Holzinger, 2016), an evolution of active learning (Settles, 2011), suggests that algorithms may request information which they recognize as missing from external agents (humans) to assist with such issues. While it is not necessary for the uncertainty algorithm to identify new classes, such "human-in-the-loop" label assignments would be useful in capitalizing on the results of self-supervised learning.

As such, we enacted a scenario wherein the uncertainty algorithm requested human-relevant labels for the new classes it produced following a period of self-supervised learning. As the algorithm created new classes, those that represented three or more samples were flagged for post-analysis. Since all of the samples shown in the experiment were from Set C, they were given the same label.

Following procedures described in sec. 5.1.1-2, the learning system was again tested on the ground training data as well as aerial test Sets A and B. The results of this experiment are summarized in Table 6. Accuracy across Set A and B objects from the ground or air did not change, which confirms that catastrophic forgetting did not occur as a result of the self-supervised learning. Additionally, when the system was evaluated on aerial test Set C images, consisting entirely of the previously unknown object class, its performance jumped from 0.0%, representing complete unknown, to 42.0%. While this value was low, it must be noted that not all of the classes created were subsequently assigned a human-relevant label. It is highly possible, if not likely, that the remainder of the test set was classified among the newly-created, but unlabeled, classes. Even so, the testing improvement resulted from exposure to only 29 total samples, and only 13 of those samples were given labels through post-analysis. This final learning experiment shows that even with only a small dataset available, our uncertainty algorithm is capable not only of continual learning but of self-supervised one-shot learning. Here, human involvement is only necessary to assign human-readable labels. The ability of the learning system to do one-shot learning by identifying and grouping new objects in a *self-supervised* manner, without catastrophic forgetting of other categories, represents a breakthrough in machine learning.

**Table 6: One-Shot Learning with Set C Objects**. Table lists system classification performances on ground data and aerial test Sets A, B, and C before and after exposure to and subsequent labeling of Set C objects. When tested with Set C objects, the learning system successfully identified almost half of the test samples with the human-given Set C identifier, improving greatly from when it could not recognize Set C at all. Performances on other test sets did not diminish.

|  | Ground Set A+B | Aerial Set A | Aerial Set B | Aerial Set C |
|---|---|---|---|---|
| **Initial Performance** | 96.7% | 92.3% | 94.1% | 0.0% |
| **After Aerial C Few-Shot Labeling** | 96.7% | 92.3% | 94.1% | 42.0% |

### 5.2 Embodied Drone Agent Experiment

The learning experiments described in sec. 5.1 verified the capability of the UML algorithm for continual learning, and were framed as a proof-of-concept for a ground-based agent learning to perform tasks after being transferred to an air-based platform. Next, we constructed a fully-integrated learning system within a 3D simulated environment and drone platform.

For this embodied agent experiment, the drone was loaded with a similar network to that trained at the start of the first learning experiment in sec. 5.1.1, trained only with ground views of objects within Sets A, B, and O. For this experiment, the detection subsystem was modified to return only vehicle-like objects (see sec. 5.2.1-3). The experiment began with the drone flying to a location 30m above the Set A intersection, where it exhibited poor initial classification performance. To demonstrate self-supervised learning, the drone was instructed to acquire object IDs by flying to an altitude that ensured accurate identification. Since accuracy is close to 100% from the ground, the drone was sent to a location approximately 2m above ground, where the spatial attention subsystem, as described in sec. 5.2.2, can use the class ID of segmented objects to define the ground truth labels and spatial location for all objects. The drone then continued its self-supervised learning by hovering at higher altitudes while observing the same scene. The drone climbed to 10m and then reached 30m in 5m increments. At each altitude, it sampled the scene and used ID information stored in its spatial memory to refine its recognition categories for all observed vehicles. To achieve the observation of the vehicles from multiple views, which supports improved recognition robustness, the drone was sent to two additional locations around the intersection (executing a 45 and 90 degrees sweeps from the initial location). Learning was limited to 20-30 frames at each altitude step and location. UML evaluates each frame and based on uncertainty measures for each object, it may accept a correct classification, refine its priors, or establish new recognition templates to accommodate significant differences in object appearance due to the drastic changes in view perspective. Figure 7 shows the object classifications before and after learning. Here, it is evident that in each case the UML algorithm appropriately learned how to classify each object.

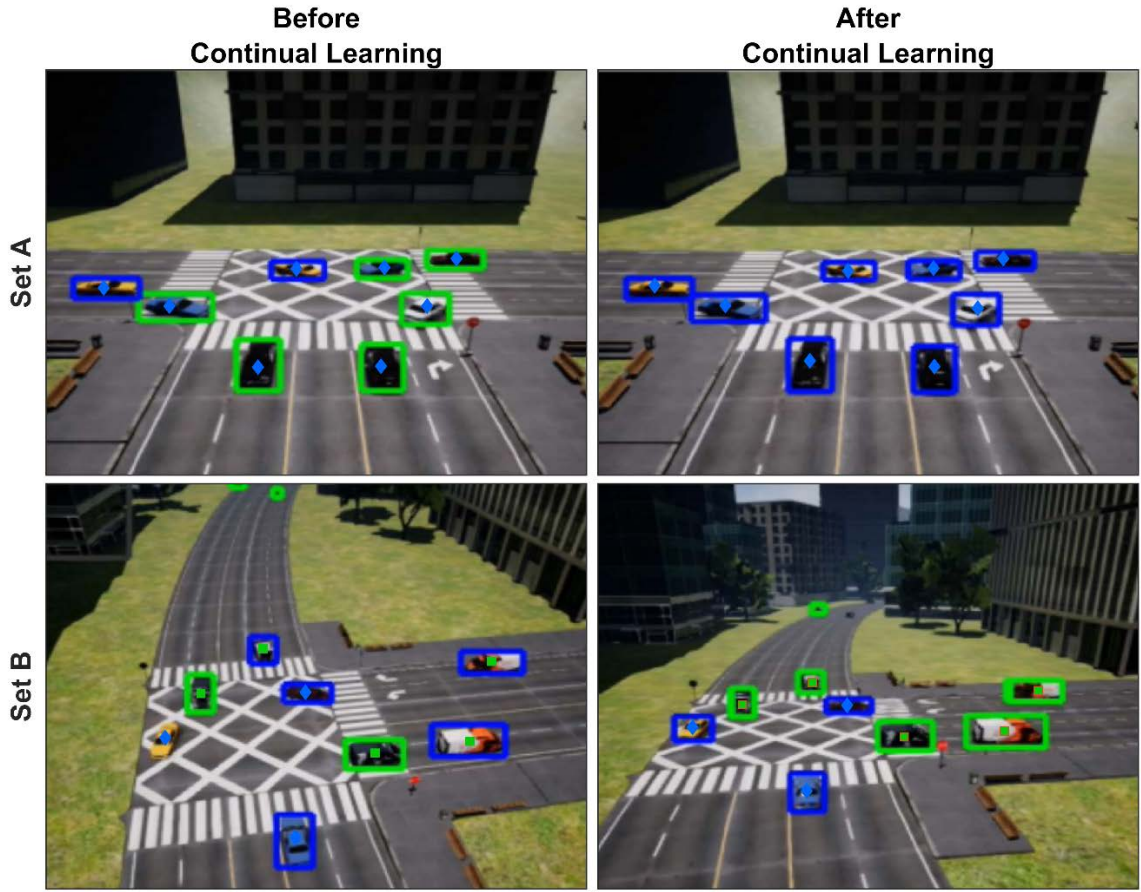

**Figure 7: Embodied System Performance.** Representative screenshots from the embodied system. Top row: Screenshots from the first intersection with sedans present before (left) and after (right) continual learning. Green boxes correspond to "van" labels, while blue boxes correspond to "sedan" labels. Shapes inside the boxes indicate the ground truth: blue diamonds = sedans, green box = van. Bottom row: Screenshots from the second intersection with vans and sedans before (left) and after (right) continual learning. In both instances, classification improves to near perfect performance after continual learning.

Following the first stage of the experiment, the drone was commanded to fly at 30m altitude towards the second intersection, which contained objects from both Sets A and B (sedans and vans). Here, the agent was able to recognize objects from Set A, but not Set B. The agent then followed the same procedure executed at the first intersection to learn Set B objects through self-supervision. Following learning at the second intersection, the drone was sent back to the Set A intersection for a final validation test. From a 20m altitude, the uncertainty algorithm correctly identified all the objects at the first intersection. Therefore, we were able to confirm that even in a real-time operating agent, the UML algorithm exhibited robustness against catastrophic forgetting.

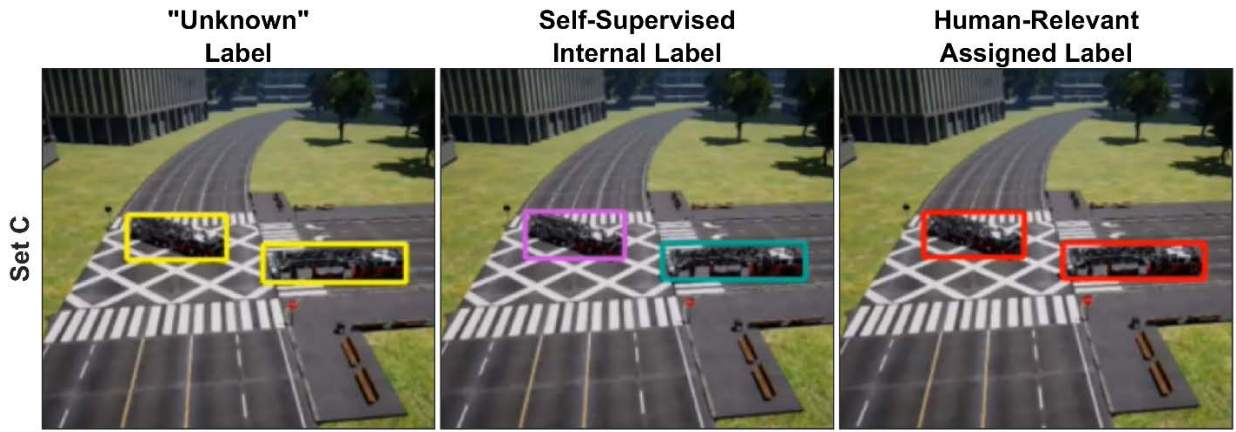

**Figure 8: Embodied System in Self-Supervised One-Shot Learning.** Screenshots showing the UML algorithm's progression through stages of One-Shot Learning. When first encountering the fire trucks, the algorithm first categorizes the novel object as "Unknown". Next, the algorithm creates *de novo* class IDs and continues to learn the representation of the new objects. Finally, a human assigns the correct label to the new classes, which combine to form the single class of "Fire Truck".

The final stage of the experiment began after the drone was sent to a third intersection with Set C objects at an altitude of 20m (see Figure 8). The architecture correctly identified the previously unseen firetrucks as unknowns, marking them with representative yellow bounding boxes. As the algorithm engaged in learning, it created a small number of new classes(magenta and green boxes). Interestingly, the two objects were largely assigned to three self-generated classes and there was little to no apparent crossover of classes between the two objects. This observation supports the algorithm's property for learning and subsequently modifying general views of object classes. Finally, emulating the process of a human-in-the-loop process for assigning human-readable labels, these new classes were manually assigned a "Firetruck" label to associate the new classes within the algorithm. With this new information (few-shot learning), the algorithm was able to gain the ability to label additional views of Set C objects (red boxes). A video of this experiment is included in the Supplementary Materials and can be accessed at https://youtu.be/yjUSdYSd0ZY.

## 6. Future Work

The experiments described here have demonstrated that the uncertainty algorithm has key capabilities to enable continual lifelong learning. However, there are aspects of the fully-integrated system that are not yet reflective of an ideal learning agent. Among them is the fact that while the system can learn, it does not include a built-in mechanism to seek out new information on which to do so. In the embodied experiment, waypoints were pre-scripted; we populated an environment with intersections that had specific objects to learn about. Outside of a simulation environment or in a more dynamic scenario, a learning agent will need to be able to determine on its own what is necessary to learn to improve its function. Reinforcement learning (Sutton & Barto, 1998) addresses this issue in a way that could easily be implemented within the learning system as developed. Reinforcement learning promotes goal-oriented behaviors by rewarding a learning agent when it successfully performs a task related to that goal. Combining rewards with different learning behaviors would incentivize a continual learning agent to explore its environment for information it can acquire. If the rewards are designed well (Kaelbling, Littman, & Moore, 1996), the learning agent may seek out information that would be directly useful for its designated tasks. In the context of our algorithm, a straight-forward extension is to reinforce uncertainty-reducing behaviors. Towards this end, we are in the process of incorporating components of reinforcement learning to provide the agent tools to intelligently explore its environment and optimize its learning.

Another aspect of the learning system that holds untapped potential relates to how a machine learning architecture interacts with the features it examines. This version of the uncertainty algorithm examines uncertainties only at the level of classification; it is assumed that the detection subsystem returns complete, clean features that perfectly reflect the nature of the detected objects. However, in a real-world scenario, this may not always be the case, as changes in weather, motion blur, and noise in the sensor among other things may degrade the quality of the captured image and the features subsequently drawn from it. ART and neurobiological mechanisms both suggest that uncertainty may be able to drive intelligent modulation of learning processes (Dasgupta et al., 2018; Hasselmo & McGaughy, 2004; Yu & Dayan, 2005). For instance, changing their relative values to direct a system's attention to those features more informative for the task at hand. We are currently examining the use of the uncertainty algorithm at earlier stages of the learning system to allow a learning agent to adapt its processing capabilities for input signals with varied characteristics and levels of quality (i.e., adapt to noisy inputs).

As a final note, it is important to keep safety in mind when designing, and ultimately building, continual, lifelong learning systems. The chief goal of learning systems like our uncertainty algorithm is to be able to take a limited starting knowledge base, expand on it, and shape it to produce a better understanding of the system's environment and improve its function. Much like with children, the ability for a learning system to form intelligent decisions/hypotheses will come with experience (exposure), but decisions early in "development" should be limited in their scope and impact until safe and effective operation is established (Varshney, 2016). Our architecture examines specific types of uncertainty that prevent it from finalizing hypotheses for which it has inadequate evidence, and how it can follow principles of interactive and active machine learning to request additional guidance as necessary. However, we are exploring additional routes to ensure the architecture protects parts of its knowledge base that should remain unchanged in order to maintain safe operation.

## 7. Conclusion

The work presented here demonstrates that our uncertainty-modulated learning architecture can enable an embodied agent, in real-time, to recognize new information in a self-supervised manner and to adjust its pre-existing knowledge base continually over long periods of time. Inspired by ART and powered by modulatory mechanisms, the system utilizes uncertainty metrics and thresholds to guide its learning in a closed-loop with its environment. Relatively simple changes to uncertainty criteria allow for multiple learning and adaptation states, including recognition of unknown objects and subsequent one-shot learning of those objects. These capabilities well surpass those of pre-existing machine learning algorithms, and through further enhancement, such as by incorporating reinforcement learning and feature adaptation, we hope to continue to contribute to new machine learning algorithms capable of robustly solving the stability-plasticity dilemma while optimally tuning their performance over their lifetimes.

## 8. Acknowledgements

This work was partially supported by the Air Force Research Laboratory (AFRL) and the Lifelong Learning Machines program by DARPA/MTO under Contract No. FA8650-18-C-7831. Any opinions, findings and conclusions or recommendations expressed in this material are those of the author(s) and do not necessarily reflect the views of AFRL and DARPA.

## 9. Declaration of Interest

The authors have the following competing interests to declare. A provisional patent for technology described herein has been filed on behalf of the Teledyne Scientific Company, Research Triangle Park, North Carolina.

Manuscript under review by Neural Networks journal